\documentclass[accepted]{uai20212} 

\usepackage{amsmath}
\usepackage{times}
\usepackage[utf8]{inputenc}
\usepackage{graphicx,subcaption}
\usepackage{color}
\usepackage{multirow}
\usepackage{amsfonts}
\usepackage{amssymb}
\usepackage{apacite}
\usepackage{caption}
\usepackage{booktabs}
\usepackage{hyperref}
\usepackage{tabularx,booktabs}
\newcolumntype{C}{>{\centering\arraybackslash}X} 
\usepackage{lipsum}
\usepackage{etoolbox}
\usepackage{tikz}
\usetikzlibrary{tikzmark}
\usetikzlibrary{calc}
\usepackage{lscape}

\usepackage{rotating}
\usepackage{bbm}
\usepackage{latexsym}
\newtheorem{definition}{Definition}

\usepackage[american]{babel}

\usepackage{natbib} 
\usepackage{mathtools} 
\usepackage{booktabs} 
\usepackage{tikz} 

\title{A Kernel Framework to Quantify a Model's Local Predictive Uncertainty under Data Distributional Shifts}

\author[ ]{\href{mailto:Rishabh Singh <rish283@ufl.edu>?Subject=Your paper}{Rishabh~Singh}}
\author[ ]{\href{mailto:Jose C. Principe <principe@cnel.ufl.edu>?Subject=Your paper}{Jose~C.~Principe}}
\affil[ ]{
    Computational NeuroEngineering Lab\break
    University of Florida\break
    Gainesville, Florida, USA
}

\affil[ ]{\href{mailto:Rishabh Singh <rish283@ufl.edu>?Subject=Your paper}{rish283@ufl.edu}, \href{mailto:Jose C. Principe <principe@cnel.ufl.edu>?Subject=Your paper}{principe@cnel.ufl.edu}}

\begin{document}
\maketitle

\begin{abstract}
 Traditional Bayesian approaches for model uncertainty quantification rely on notoriously difficult processes of marginalization over each network parameter to estimate its probability density function (PDF). Our hypothesis is that internal layer outputs of a trained neural network contain all of the information related to both its mapping function (quantified by its weights) as well as the input data distribution. We therefore propose a framework for predictive uncertainty quantification of a trained neural network that explicitly estimates the PDF of its raw prediction space (before activation), $p(\tilde{y}|x, w)$, which we refer to as the model PDF, in a Gaussian reproducing kernel Hilbert space (RKHS). The Gaussian RKHS provides a \textit{localized} density estimate of $p(\tilde{y}|x, w)$, which further enables us to utilize gradient based formulations of quantum physics to decompose the model PDF in terms of multiple local uncertainty moments that provide much greater resolution of the PDF than the central moments characterized by Bayesian methods. This provides the framework with a better ability to detect distributional shifts in test data away from the training data PDF learned by the model. We evaluate the framework against existing uncertainty quantification methods on benchmark datasets that have been corrupted using common perturbation techniques. The kernel framework is observed to provide model uncertainty estimates with much greater precision based on the ability to detect model prediction errors.
\end{abstract}

\section{Introduction}
Deep neural network (DNN) models have become the predominant choice for pattern representation in a wide variety of machine learning applications due to their remarkable performance advantages in the presence of large amount data \citep{lec}. The increased adoption of DNNs in safety critical and high stake problems such as medical diagnosis, chemical plant control, defense systems and autonomous driving has led to growing concerns within the research community on the \textit{performance trustworthiness} of such models \citep{kend, lund}. This becomes particularly imperative in situations involving data distributional shifts or the presence of out of distribution data (OOD) during testing towards which the model may lack robustness due to poor choice of training parameters or lack of sufficiently labeled training data, especially since machine learning algorithms do not have extensive prior information like humans to deal with such situations \citep{amod}. An important way through which trust in the performance of machine learning algorithms (particularly DNNs) can be established is through accurate techniques of \textit{predictive uncertainty quantification} of models that allow practitioners to determine how much they should rely on model predictions.\par

\begin{figure*}[!t]
    \centering\includegraphics[scale = 0.3]{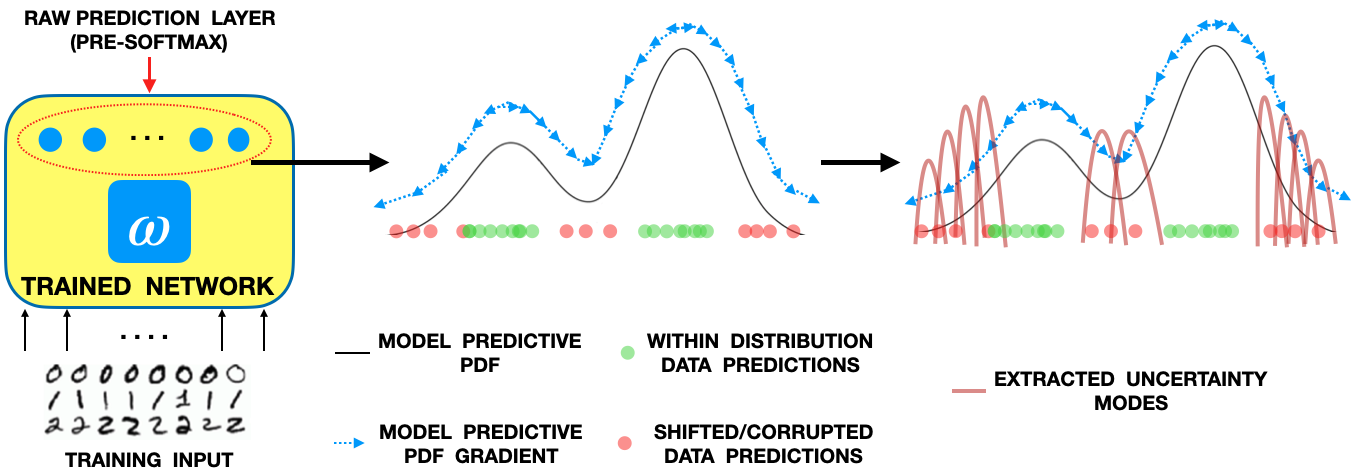}
  \caption{Kernel Uncertainty Framework: (i) Estimation of model predictive PDF. (ii) Quantification of local PDF gradient flow. (iii) Extraction of uncertainty moments, which provide multi-scale resolution of corrupted input predictions that typically lie at PDF inflection and tail regions, away from dense PDF regions which reflect most frequent training data.}
\end{figure*}

Existing methods for model uncertainty quantification can be broadly classified into Bayesian \citep{mack, neal, bishop} and non-Bayesian techniques \citep{tib, osb, pearce}. Bayesian methods involve inference of probability distributions over model weights. Early work in this domain includes methods such as Markov-chain Monte Carlo based Bayesian neural networks \citep{bishop}, Hamiltonian Monte Carlo \citep{neal} and Laplacian approximation \citep{mack}. Although such methods provide mathematically grounded approaches towards probabilistic inference by marginalizing over model weights, their applicability towards large datasets and model architectures becomes restrictive due to huge computational overheads involved. To mitigate this problem, faster variational inference based methods were developed later on to increase the training speed and efficiency of Bayesian neural networks \citep{graves, jord, hoff}. However, such methods still suffer from high dimensionality and high complexity of weight associations making them infeasible for efficient learning of parameter dependencies \citep{projbnn}. Non-Bayesian techniques include ensemble based methods that aggregate outputs of multiple differently initialized models to give probabilistic outputs. A significant work in this line is that of \citet{laks}, where authors use deep ensemble neural networks to quantify output uncertainty. A work that recently gained a lot of popularity is that by \citet{gal} who suggest to use Monte Carlo Dropout during model testing as an efficient way of extracting output uncertainty information. However, the increasing requirements of number of parallel trained models and the number of stochastic forward passes in ensemble and MC dropout methods respectively gives rise to concerns regarding their tractability in modern applications.\par

Despite recent advances in the field, current methods lack in their ability to provide high resolution quantification of model uncertainty that explains prediction results for single inputs. This is due to their ignorance of the local model structure and its dependencies. Our work focuses on the density and local fit criteria \citep{leo} according to which a prediction at any point $x$ is reliable/certain only if there is a significant local density of samples around $x$, or a prediction $y$ is reliable only if there is a significant local density of predictions around (or similar to) $y$ made by the model. To quantify the model space according to this criteria, a rich localized representation of the model predictive PDF is needed. Our conjecture is that the Gaussian reproducing kernel Hilbert space (RKHS) would be the most ideal tool for such a representation because of its universal data characterization properties through a \textit{localized} functional representation of the data space \citep{aron}. Moreover, the RKHS also provides a natural coordinate-free representation for statistical modeling \citep{parzen1970}.\par

\section{Problem Formulation}
We define the problem setup as follows. Let us assume that $x \in \mathbb{R}^d$ represents $d$-dimensional input features and $y \in \{1...k\}$ represents the target labels for $k$-class classification. We represent the training dataset with $\mathbb{D}$ which consists of $N$ i.i.d samples so that $\mathbb{D} = {\{(x_{tr}, y_{tr})\}}_{tr=1}^N$. We further denote the unknown probability density function observed through samples of $\mathbb{D}$ (i.e. the true distribution of the classification problem) as $p^*(x,y)$. We assume a neural network parameterized by $w$ and trained on $\mathbb{D}$ to model the conditional distribution $p(y|x, \omega)$, where $\omega$ is learnt from the training set. Let us further assume the training input samples to be governed by an unknown distribution $p(x|\lambda)$.\par

Our goal is to quantify the predictive uncertainty of the trained neural network model on a test set whose governing PDF has undergone a shift from $p(x|\lambda)$ to become another unknown distribution $p(x|\gamma)$, while the target conditional distribution, $p(y|x)$, remains the same. This is also known as \textit{covariate shift} \citep{sug}. In order to accomplish this, we rely on modal decomposition of the dynamics of probability density function (PDF) associated with the raw prediction space of the model (which we henceforth also refer to as the model PDF) given by $p(\tilde{y}|x, \omega)$, where $\tilde{y} \in \mathbb{R}^k$ is the $k$-dimensional raw prediction vector (output before softmax) corresponding to the input $x \in \mathbb{R}^d$ of a trained neural network model parameterized by weights $\omega$. We then desire to quantify the gradient of the model PDF at each point in its space as a combination of successive orthogonal uncertainty moment functionals ($h^1, h^2, ..., h^t$) given by $\Delta{p_{\tilde{y}|x, \omega}}(x_i) = h^1_i + h^2_i + ... + h^t_i$, so that the moments systematically quantify the regions of the model PDF space exhibiting varying degrees of change in the local density, which fall in the inflection and tail regions of the model PDF that have high associated uncertainties. Essentially, we want to estimate the confidence of a model in its categorical prediction $y$ on the basis of the local density around its corresponding value in the space of $\tilde{y}$. This is a non-trivial task and it is well known that softmax function (for classification), gives unreasonably high probabilities of predictions whose corresponding values in the space $\tilde{y}$ fall outside of the model PDF \citep{gal}.

\section{Contributions}
We propose to utilize a recently introduced kernel framework for uncertainty quantification that has shown promising preliminary results for particular applications in time series analysis \citep{stm} as well as model uncertainty quantification for regression problems \citep{stw}. Here, authors utilize the Gaussian reproducing kernel Hilbert space (RKHS) to infer the implicit PDF of the model and quantify its gradient flow in the local space of its prediction points in order to obtain uncertainty scores. We think that this framework would be well suited for predictive uncertainty quantification problems involving distributional shifts in the test-set because of its high emphasis on the \textit{local PDF behavior} of the model (not seen in traditional methods) to quantify its uncertainty. Our implementation of the framework, depicted in fig. 1, first involves a local estimation of the model PDF, $p(\tilde{y}|x, \omega)$. This is done through a functional mapping of $\tilde{y}_{tr}$ (raw prediction output of the training input data) in the Gaussian RKHS by a metric called the \textit{information potential field} (IPF) which provides a probability measure at every point in the entire space of $\tilde{y}$ induced by $\tilde{y}_{tr}$. This subsequently allows one to utilize Laplacian operator based formulations in physics to quantify the local gradient flow of the model PDF at every point of $\tilde{y}$. Specifically, we leverage the Schr\"odinger's formulation which expresses the local gradient flow of the IPF as a moment decomposition problem with the higher order moments signifying regions close to the inflection and tail regions of the PDF where its gradient flow is maximum (maximum uncertainty). This therefore makes the uncertainty framework tailored towards applications where there are discrepancies of varying nature between the distributions of the training data (model's learnt PDF) and the test data, either due to a distortion or a presence of out-of-distribution (OOD) test data.\par

Such a framework presents a simpler and a more efficient alternative to traditional Bayesian viewpoint of PDF inference and uncertainty quantification that relies on the notoriously difficult process of parameter marginalization through variational inference or Monte Carlo based methods. The mathematical properties of the Gaussian RKHS, specifically the kernel trick and the kernel mean embedding theory, ensures that it is a universal PDF estimator of \textit{any} function. Furthermore, PDF estimation is done incrementally in the RKHS through the accumulation of Gaussian bumps at individual points in space. This enables the use of local gradient quantifying moment based formulations (typically used in describing the dynamics of systems in physics and spectral theory) to provide a high resolution and multi-scale description of \textit{PDF dynamics} at every point in space with higher order moments quantifying regions of high PDF gradient (high uncertainty). We therefore posit that the kernel framework provides uncertainty quantification with much greater accuracy and precision than Bayesian methods that are limited to estimating the central moments of the PDF. Moreover, the proposed framework is able to provide a single-shot estimation of model uncertainty at each test instance thereby offering practical advantages over ensemble and Monte Carlo methods that rely on multiple model realizations and multiple forward stochastic runs respectively.

\section{Kernel Uncertainty Framework}
The implementation details of the uncertainty framework can be found in \citep{stw}. We highlight the details of key steps in terms of our implementation (fig. 1) as follows.

\subsection{Functional Description of Model PDF}
The foundation of utilizing the Gaussian RKHS for PDF estimation comes from the kernel mean embedding theory \citep{embor, emb} which allows one to non-parametrically and universally quantify a data distribution from the input space as an element of its associated RKHS given by $\phi(\mathbb{P}) = \mu_{\mathbb{P}} = \int k(x, .)d\mathbb{P}(x)$. Hence it's a functional representation of the data PDF induced by points $x$. It has several useful properties, one of which is the property of injectivity for characteristic kernels, meaning that $\mu_\mathbb{P} = \mu_\mathbb{T}$ if and only if $\mathbb{P} = \mathbb{T}$. This therefore allows for unique characterizations of data/model PDF. In practice, one does not have prior information regarding the specific nature of the data PDF, $\mathbb{P}$. Hence we typically rely on the unbiased empirical estimate of the KME given by $\hat\mu = \frac{1}{n}\sum_{t=1}^{n}k(x_t,.)$, which converges to $\mu$ for $n \rightarrow \infty$ in accordance with the law of large numbers.

\paragraph{Information Potential Field} A quantity that evaluates the KME in the functional space of projected samples is the information potential (IP) \citep{ux}, which is the empirical mean value of the PDF over the space of samples in the RKHS. The evaluated KME also appears in Renyi's quadratic entropy formula \citep{r2}, $H_2(X) = -log\int{p(x)^2 dx} = -log\Psi(X)$, as the argument of the logarithm denoted by $\Psi(X)$. Given experimental data $x_i, i = 1, ..., N$,  $\Psi(X)$ can be evaluated using a Gaussian kernel $G$ of width $\sigma$ as $\Psi(X) = \int p(x)^2 dx = \int\big(\frac{1}{N}\sum_{i=1}^{N}G_{\sigma}(x-x_i)\big)^2dx = \frac{1}{N^2}\int\sum_{i=1}^{N}\sum_{j=1}^{N} G_{\sigma/\sqrt{2}}(x_j - x_i)$. The information potential is mean value of kernel mean embedding (created by a particular set of samples $x_i$) evaluated at a set of points $x_j$, where $j \in {1...}N$. Its name originates from the physical interpretation of $\Psi(X)$ if we think of the projected samples as particles in a potential field. One can further express the information potential as $\Psi(X) = \frac{1}{N}\sum\limits_{j=1}^{N}\psi(x_j)$, where $\psi(x) = \frac{1}{N}\sum\limits_{i=1}^{N}G(x - x_i)$ represents the field due to the addition of Gaussians centered at each sample. We term $\psi(x)$ as the \textit{information potential field} (IPF) which is essentially the functional equivalent of a probability measure in the RKHS, i.e. it gives the probability at each point of space induced by the entire dataset. The IPF is therefore the general PDF estimator of the framework. Being an evaluation of KME, it ensures unbiased and efficient PDF estimation, with convergence independent of data dimensionality \citep{emb}. For our implementation, we compute the  model PDF at any particular raw test prediction $\tilde{y_k}$ using the IPF as follows.
\begin{equation}
    \psi(\tilde{y_k}) = \frac{1}{N}\sum\limits_{i=1}^{N}G(\tilde{y_i} - \tilde{y_k})
    \label{ipf}
\end{equation}
where $\tilde{y_i}$ denotes the $i^{th}$ training data prediction by the trained network ($N$ being the total number of training data).

\subsection{Schr\"odinger's Moment Decomposition Problem over the IPF}
 The localized structure of the IPF enables one to further utilize local operators that induces anisotropy in the IPF and transforms it into a field governed by \textit{local interactions} between the Gaussian bumps and dependent upon the local sample density. It becomes possible to employ a local differential operator (such as the Laplacian) over the IPF to formulate a decomposition in the Gaussian RKHS, which can be an indicator of uncertainty in the particular local region of space where IPF is computed. A popular formulation that leverages the Laplacian gradient operator do quantify dynamics of systems in physics is the Schr\"odinger's equation, a Laplacian operator based differential equation that operates on the \textit{local space} of the system and decomposes it as a high resolution composition of wave-function moments that quantify uncertainty at every point of the system space. This formulation therefore fits very well into a data analysis goal of  quantifying uncertainty and the structure of the RKHS allows for it to be implemented in the space of data.\par 
 
 To this end, \citet{stw} utilize a ``data-equivalent'' Schr\"odinger formulation over the IPF given by $H\psi(x) = E\psi(x)$ \citep{prin}. Here, $H$ denotes the Hamiltonian over $\psi(x)$, and $E$ is the total energy over the sample space. All the physical constants have been lumped on the only variable in the Gaussian RKHS, the size of the kernel. Similar to a general quantum system, the Hamiltonian is constructed using two operators: the kinetic and potential energy operators. The kinetic energy operator here consists of a Laplacian function over the IPF, i.e. $-\frac{\sigma^2}{2}\nabla^2\psi(x)$, which essentially quantifies local sample density in our context. We refer to the potential energy operator denoted by $V_s$ as the \textit{quantum information potential field} (QIPF). The overall formulation becomes an Eigenvalue problem over the data PDF and is given as 
$H\psi(x) = \bigg(-\frac{\sigma^2}{2}\nabla^2 + V_s(x)\bigg)\psi(x) = E\psi(x)$.
Here, the IPF, $\psi(x)$, is the probability measure and the equivalent of the wave-function (seen in general quantum systems) that is being decomposed by the Schr\"odinger's formulae. Note that $\psi(x)$ is the eigenfunction of $H$ and $E$ is the lowest eigenvalue of the operator, which corresponds to the ground state. Since, by convention in quantum systems, the probability measure is defined as the square of the wave-function ($p(x) = |\psi(x)|^2$), we rescale $\psi(x)$ in $H\psi(x)$ formulation above, to be the square root of the IPF, i.e. $\psi(x) = \sqrt{\frac{1}{N}\sum\limits_{i=1}^{N}G_{\sigma}(x-x_i)}$. We can rearrange the terms of $H\psi(x)$ to obtain the formulation of the QIPF as $V_s(x) = E + \frac{\sigma^2/2\nabla^2\psi(x)}{\psi(x)}$. To determine the value of the QIPF ($V_s(x)$) uniquely, we require that $min(V_s(x)) = 0$, which makes $E = -min\frac{\sigma^2/2\nabla^2\psi(x)}{\psi(x)}$.\par

Equivalently, we define an eigenvalue problem over model PDF as follows.

\begin{equation}
V_s(\tilde{y}) = E + \frac{\sigma^2/2\nabla^2\psi(\tilde{y})}{\psi(\tilde{y})}
\label{sf}
\end{equation}

where, $E = -min\frac{\sigma^2/2\nabla^2\psi(\tilde{y})}{\psi(\tilde{y})}$. Thus we have defined a Schr\"odinger based moment decomposition problem (QIPF) over the model PDF based on its local gradient. Given the nature of its Laplacian based formulaiton, we expect the model QIPF to exhibit its minima in locally dense regions of the model PDF space, while increasing quadratically outside of the model PDF region, thereby providing a clustering of the uncertain regions in the model PDF space.

\begin{figure*}[!t]
\begin{subfigure}{0.33\linewidth}
    \centering\includegraphics[scale = 0.15]{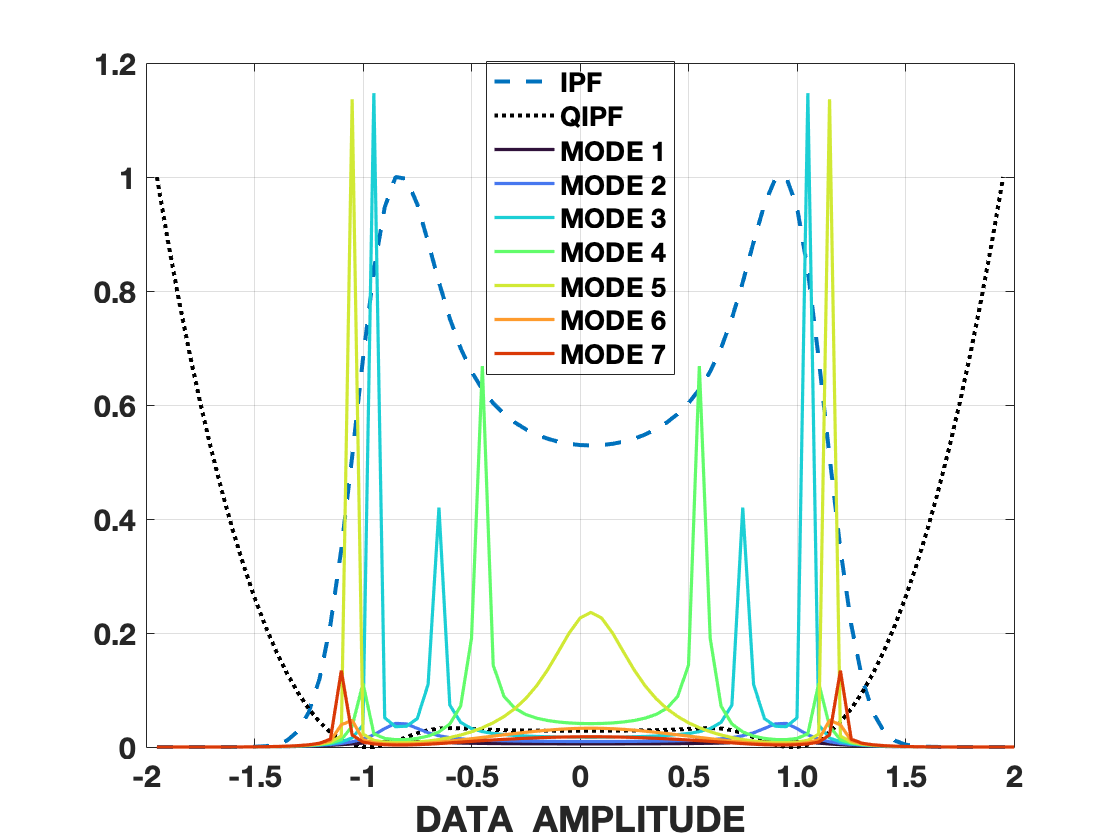}
    \caption{Kernel width:0.15}
  \end{subfigure}
  \begin{subfigure}{0.33\linewidth}
    \centering\includegraphics[scale = 0.15]{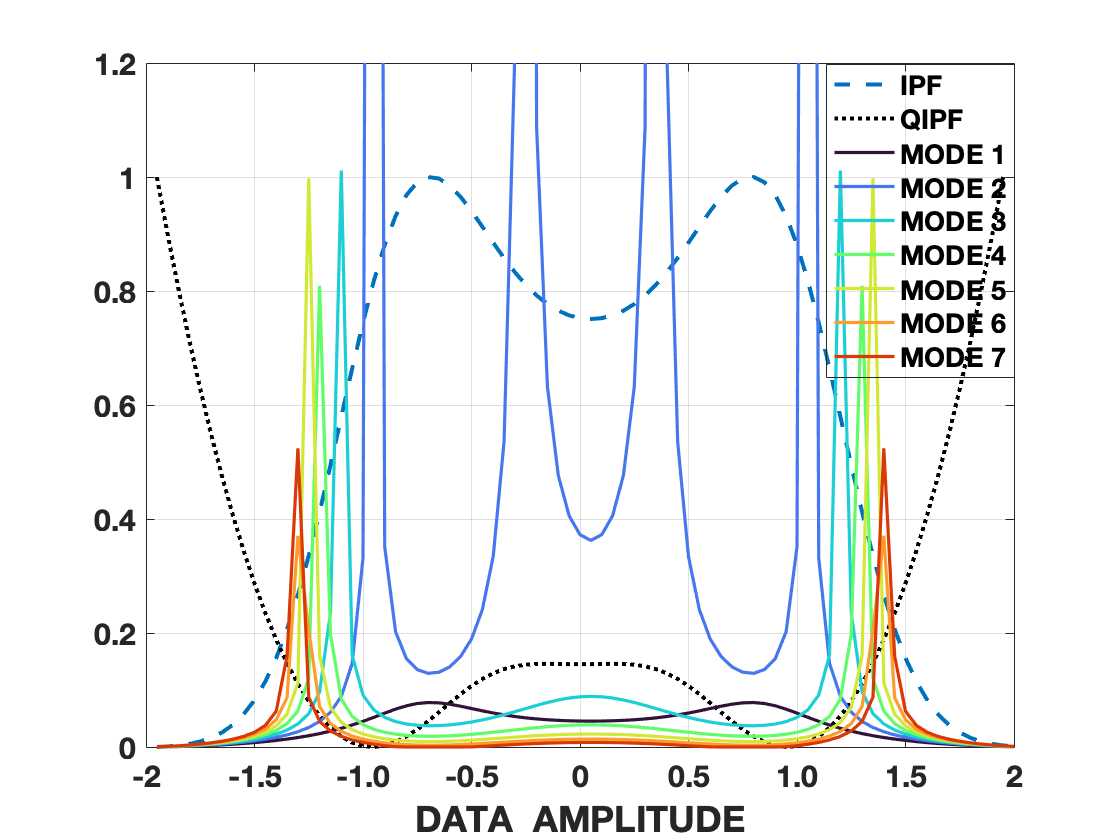}
    \caption{Kernel width:0.3}
  \end{subfigure}
  \begin{subfigure}{0.33\linewidth}
    \centering\includegraphics[scale = 0.15]{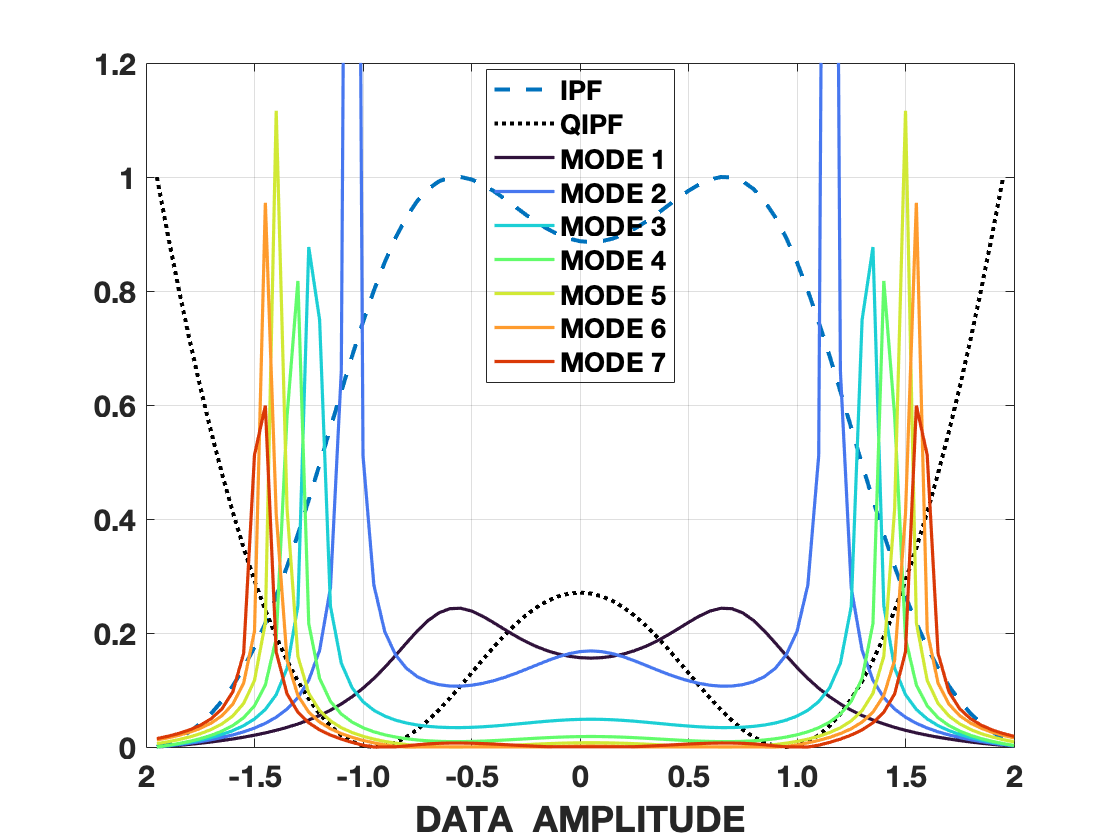}
    \caption{Kernel width:0.5}
  \end{subfigure}
  \caption{Analysis of sine-wave IPF (dashed), QIPF (dotted) and QIPF modes (solid lines) in the space of data using different kernel widths. Inflection and tail regions of the PDF are seen to be efficiently quantified by QIPF modes.}
  \label{space}
\end{figure*}

 \subsection{Definition of Model Predictive Uncertainty}
Having formulated a moment decomposition problem over the space of model PDF, we formally define model predictive uncertainty as follows.
 \begin{definition}[Model Predictive Uncertainty]
     Given a model trained on $\mathbb{D} = \{(x_{tr}, y_{tr})\}_{tr=1}^N$, the uncertainty associated with its label prediction vector $y_k$ is the decomposition of the quantum information potential field at $\tilde{y}_k$ given by $V_s(\tilde{y}_k) = V_s^1(\tilde{y}_k) + V_s^2(\tilde{y}_k) +...+ V_s^m(\tilde{y}_k)$, where $\tilde{y}_k$ is the prediction vector corresponding to $y_k$ in the model PDF space (raw prediction space before activation) and $V_s^1(\tilde{y}_k), V_s^2(\tilde{y}_k),..., V_s^m(\tilde{y}_k)$ are $m$ extracted moments of $V_s(\tilde{y}_k)$ at the point $\tilde{y}_k$.
 \end{definition}
 
 In order to extract the QIPF moments represented by $V_s^p(\tilde{y}_k)$ where $p \in {0...m}$, we follow the same approach as \citet{stw} and project the wave-function associated with the model IPF, $\psi(\tilde{y}_k) = \sqrt{\frac{1}{N}\sum\limits_{i=1}^{N}G(\tilde{y}_i - \tilde{y}_k)}$ (assumed to be in the ground state), into successive Hermite polynomial spaces given by $\psi_p(\tilde{y}) = H^*_{p}(\psi(\tilde{y}_k))$, where $H^*_p$ denotes the normalized $p^{th}$ order Hermite polynomial. This leads to the evaluation of the higher order corresponding QIPF moments as follows.
 
 	\begin{equation}
	\begin{aligned}
		V_s^p(\tilde{y}_k) = E_p + \frac{\sigma^2/2\nabla^2H^*_p(\psi(\tilde{y}_k))}{H^*_p(\psi(\tilde{y}_k))}\\\\ = E_p + \frac{\sigma^2/2\nabla^2\psi_p(\tilde{y}_k)}{\psi_p(\tilde{y}_k)}
		\label{vs}
	\end{aligned}
	\end{equation}
	where $p$ denotes the moment order number and $E_p$ denotes the corresponding eigenvalues of the various modes and is given by
	\begin{equation}
		E_p = -min\frac{\sigma^2/2\nabla^2\psi_p(\tilde{y}_k)}{\psi_p(\tilde{y}_k)}
		\label{ek}
	\end{equation}
The extracted modes of the QIPF, denoted by $V_s^p(\tilde{y}_k)$, are thus the functionals representing the different moments of uncertainty (moment order denoted by $p$) at any point $\tilde{y}_k$ in the model PDF space.\par

This follows the approach of the solution of the Schr\"odinger's equation representing a particular system in physics (quantum harmonic oscillator, for instance) that yields successive Eigenstates of the wave-function (quantifying multi-modal uncertainties of the system) that are related to each other through successive Hermite polynomials.

 \subsection{Illustrative Example}
As a pedagogical example, we implement the QIPF decomposition framework on a simple 50 Hz sine-wave signal (with mean as zero and peaking at $\pm 1$) to observe how the behaviour of uncertainty moments in the signal's data space. We computed the values of the IPF, QIPF and the first 8 modes at each point in the data space in the range (-2,2). This is shown in fig. 2 for three different kernel widths (0.15, 0.3 and 0.5). Interestingly, there is information about the sine-wave outside of the range spanned by its samples $(-1, 1)$. One can observe here how the IPF quantifies the bi-modal PDF of the sine-wave, with the peaks signifying regions of highest sample density. The QIPF, which quantifies the PDF gradient, has the lowest values at the two peak locations and slightly increases at the valley between the peaks (i.e. the mean region, where there is considerable PDF gradient). It increases significantly from the tail regions to outside of the signal's PDF, where the local sample density constantly decreases. Among the extracted modes of the QIPF, the lower order modes can be seen to peak at the vicinity of the PDF peaks where the PDF gradient starts to increase. The higher order modes can be seen to successively cluster further away at the tail regions of the PDF where the gradient is maximum, thus signifying the most locally uncertain regions of the data space. For lower kernel widths (fig. 2a, 2b), some middle order modes can also be seen to peak at signal's mean region because of its relatively high gradient and low sample density as compared to the peaks. At the higher kernel width of 0.5, since the IPF lumps both the peaks in one single cluster, the density becomes more uniform from the peaks to the mean, thus resulting in no modes peaking at the mean. Hence it can be seen how the QIPF moments are highly sensitive towards the local behavior of the signal PDF and provide a high resolution quantification of uncertainty based on sample density and gradient. While the QIPF itself does not exhibit such a high local sensitivity towards the PDF behavior like its extracted moments do, it can be seen as an appropriate measure for quantifying out-of-distribution regions which the modes do not quantify as well due to absence of samples.


 \section{Methods and Metrics}
 For comparisons, we follow the same experimental setup as done by \citet{laks2}, where authors perform an extensive evaluation of the different predictive UQ methods on several benchmark classification datasets under test-set corruption. We use the same established methods as comparison benchmarks that, much like the QIPF framework, use no prior information about the nature or intensity of test-set corruption to quantify model uncertainty and hence fall into the same category of methods. These methods are listed as follows.
 \begin{itemize}
 	\item \textbf{MC Dropout:} Monte Carlo implementation of dropout during model testing \citep{gal, sriv2}.
 	\item \textbf{MC Dropout LL:} MC dropout implemented only over the last model layer \citep{riq}.
 	\item \textbf{SVI:} Stochastic Variational Bayesian Inference \citep{graves, blun}.
 	\item \textbf{SVI LL:} Stochastic Variational Bayesian Inference applied only to the last model layer \citep{riq}.
 	\item \textbf{Ensemble:} Ensembles of T (set as 10 here) networks independently trained with different random initializations on the datasets \citep{laks}.
 \end{itemize}

To evaluate the performance of the different UQ methods, we differ from \citet{laks2}, where authors have used various calibration metrics to compare them. Instead, we evaluate the UQ methods based on their ability to detect classification errors of the model in real time when fed with corrupted test data. This, in our opinion, is a more practical method of evaluating UQ techniques since it directly measures the \textit{online (sample-by-sample)} \textit{trustworthiness} associated with prediction models. We measure this ability both in terms of areas under ROC and precision-recall curves associated with the error detection problem. Additionally, we also evaluate the point-biserial correlation coefficient (used for measuring correlation between a dichotomous and a continuous variable) between the uncertainty quantified by the above techniques (continuous variables) and the model prediction errors (dichotomous variable).

\begin{figure*}[!t]
\begin{subfigure}{0.24\linewidth}
    \centering\includegraphics[scale = 0.18]{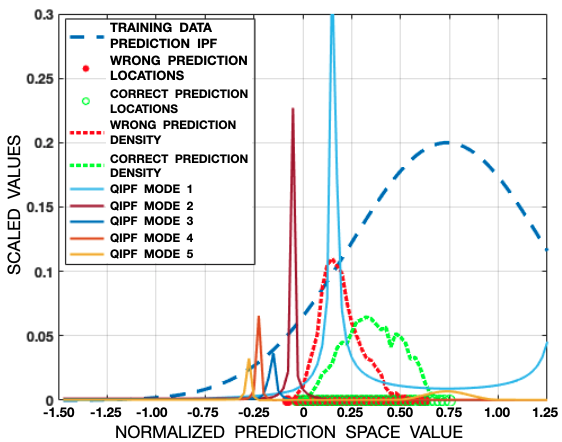}
    \caption\centering{Predictive IPF, test prediction densities, QIPF modes}
  \end{subfigure}
  \begin{subfigure}{0.5\linewidth}
    \centering\includegraphics[scale = 0.24]{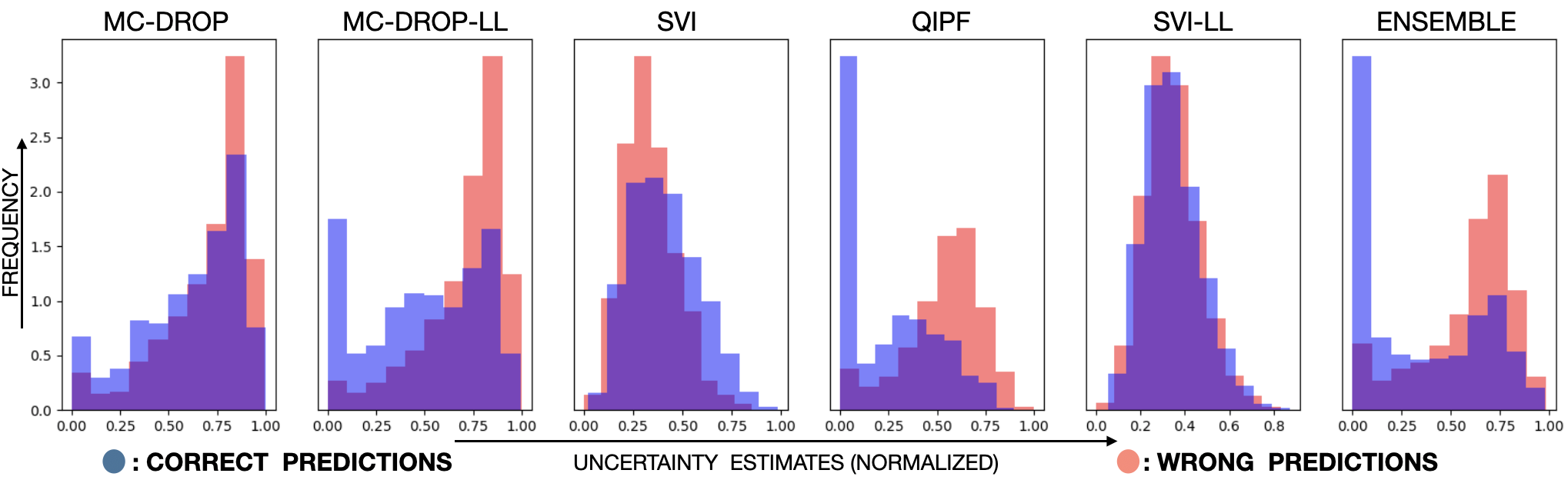}
    \caption\centering{Histogram of Uncertainty Scores}
  \end{subfigure}
  \begin{subfigure}{0.252\linewidth}
    \centering\includegraphics[scale = 0.18]{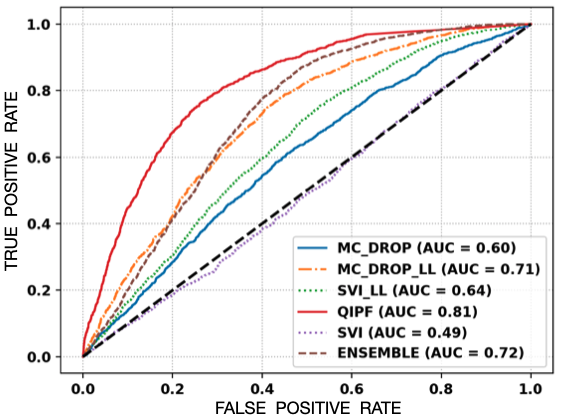}
    \caption\centering{ROC Curves}
  \end{subfigure}
    \begin{subfigure}{0.24\linewidth}
    \centering\includegraphics[scale = 0.18]{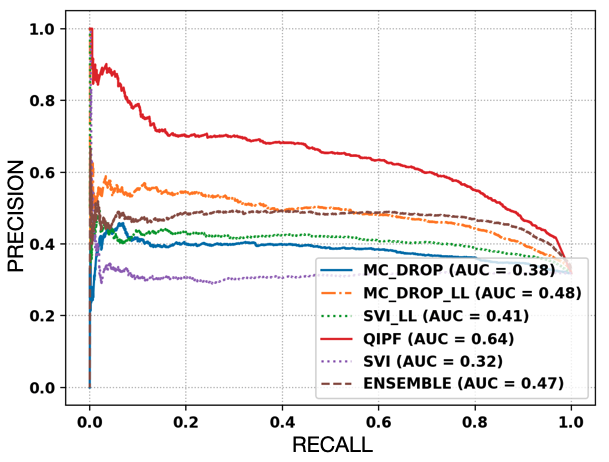}
    \caption\centering{Precision-recall curves}
  \end{subfigure}
    \begin{subfigure}{0.24\linewidth}
    \centering\includegraphics[scale = 0.18]{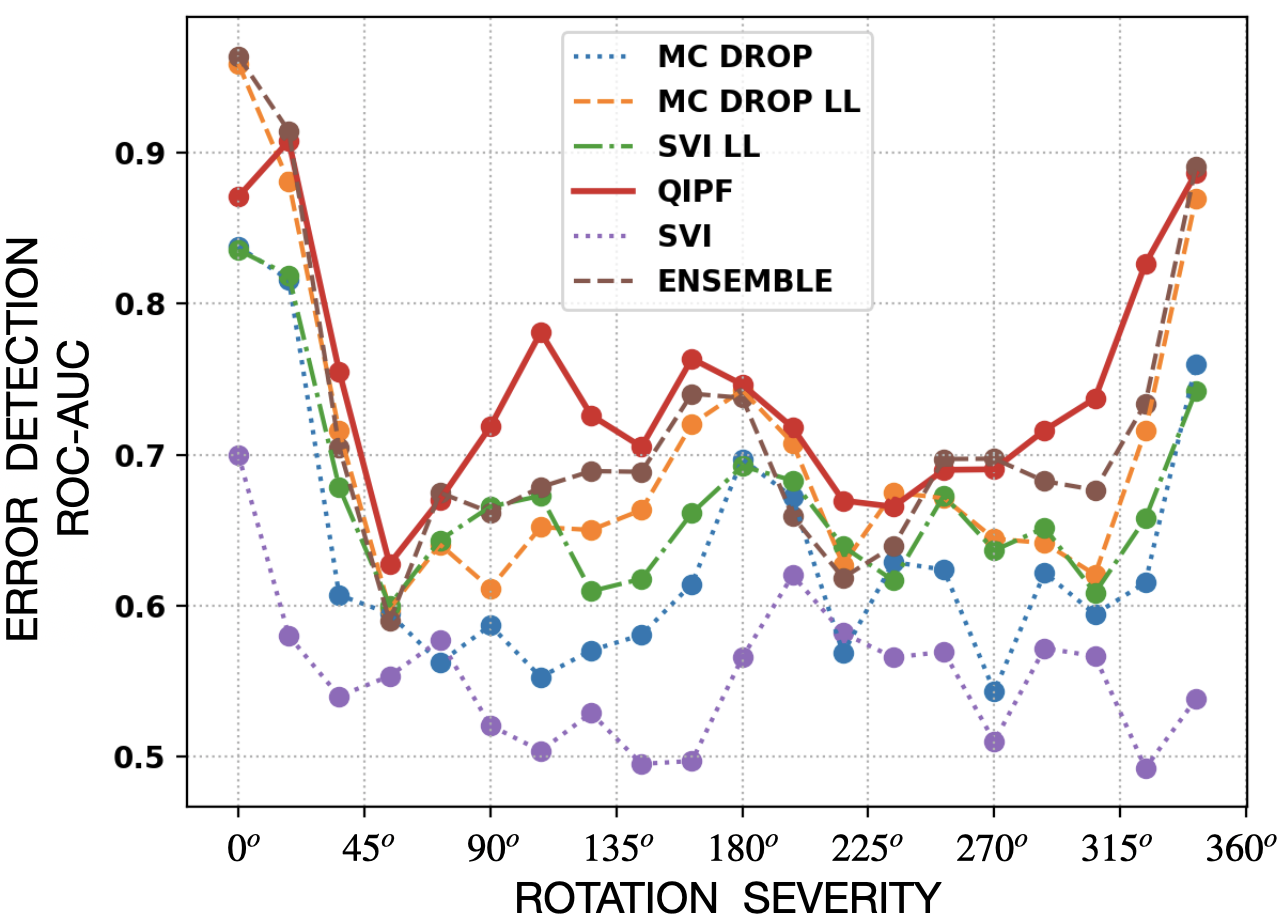}
    \caption\centering{ROC-AUC vs noise severity}
  \end{subfigure}
    \begin{subfigure}{0.24\linewidth}
    \centering\includegraphics[scale = 0.18]{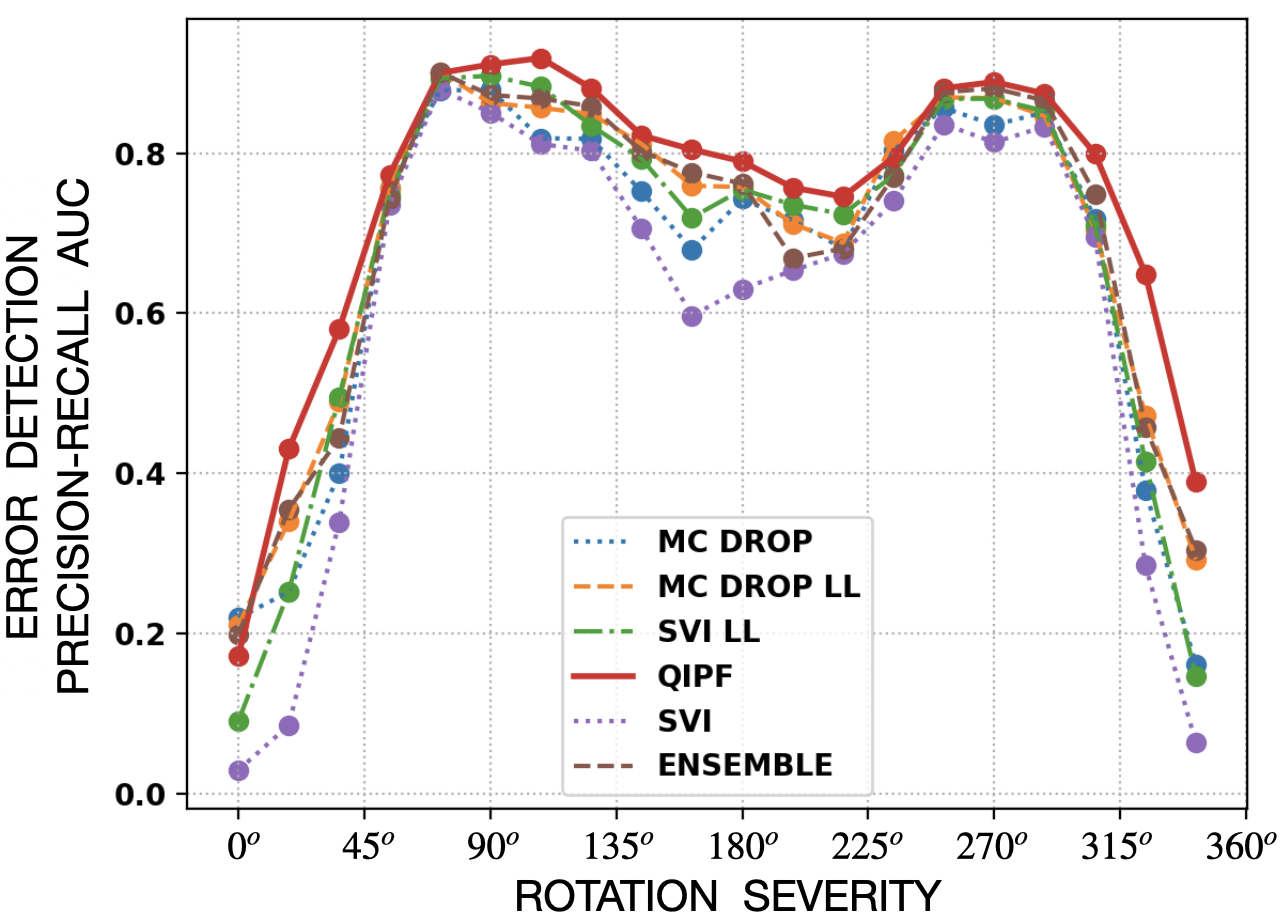}
    \caption\centering{PR-AUC vs noise severity}
  \end{subfigure}
  \begin{subfigure}{0.24\linewidth}
    \centering\includegraphics[scale = 0.18]{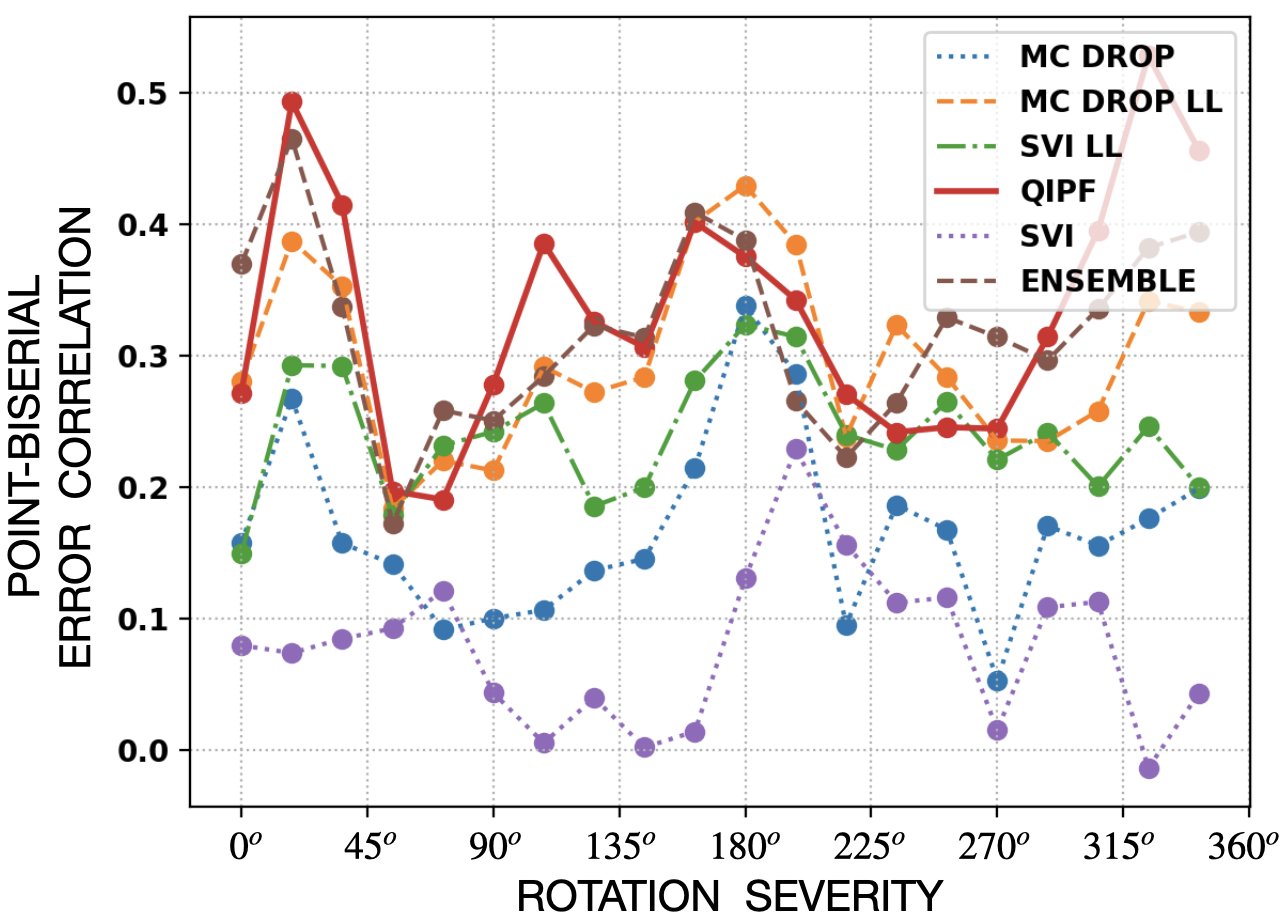}
    \caption\centering{Error Corr. vs noise severity}
  \end{subfigure}
    \caption{MNIST illustrative example: (a) shows how the QIPF modes discriminatively quantify the prediction space, (b) shows the histograms of uncertainty estimates where QIPF can be seen to achieve better class-separation between correct and wrong predictions. (c) and (d) show error detection ROC and precision-recall curves, (e), (f) and (g) show the graphs of ROC-AUC, PR-AUC and PT-Biserial Corr. vs noise internsity. QIPF framework can be seen to have better performance.}
  \label{mni}
  \end{figure*}

 \begin{figure*}[!t]
   \begin{subfigure}{0.33\linewidth}
    \centering\includegraphics[scale = 0.2]{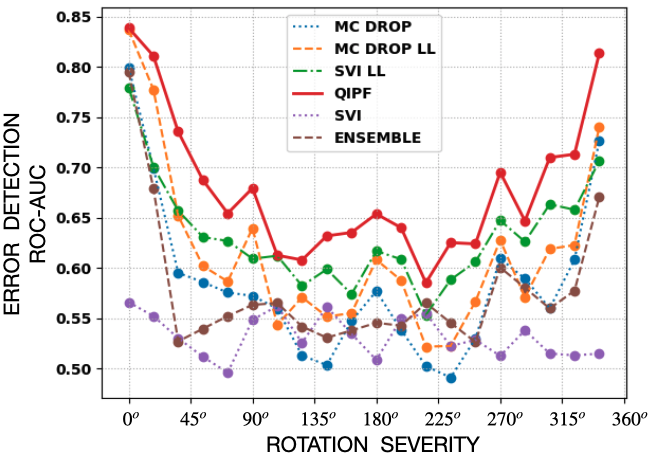}
    \caption{ROC-AUC vs Rotation Severity}
  \end{subfigure}
    \begin{subfigure}{0.33\linewidth}
    \centering\includegraphics[scale = 0.2]{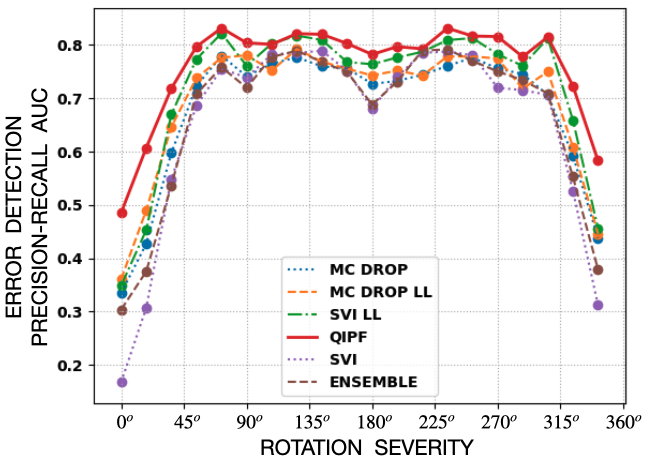}
    \caption{PR-AUC vs Rotation Severity}
  \end{subfigure}
  \begin{subfigure}{0.33\linewidth}
    \centering\includegraphics[scale = 0.2]{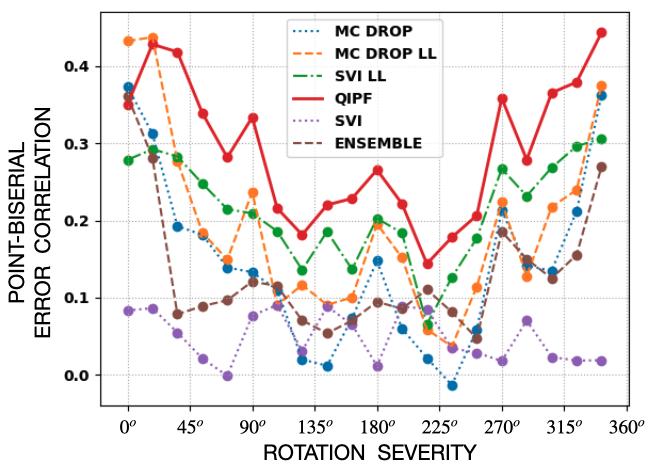}
    \caption{Error Corr. vs Rotation Severity}
  \end{subfigure}
    \begin{subfigure}{0.33\linewidth}
    \centering\includegraphics[scale = 0.2]{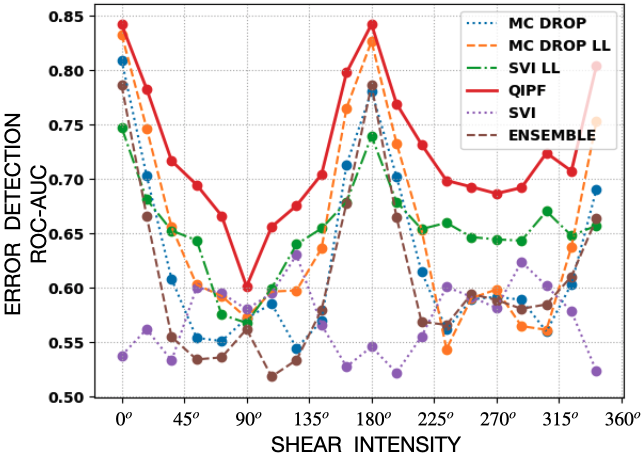}
    \caption{ROC-AUC vs Shear Intensity}
  \end{subfigure}
    \begin{subfigure}{0.33\linewidth}
    \centering\includegraphics[scale = 0.2]{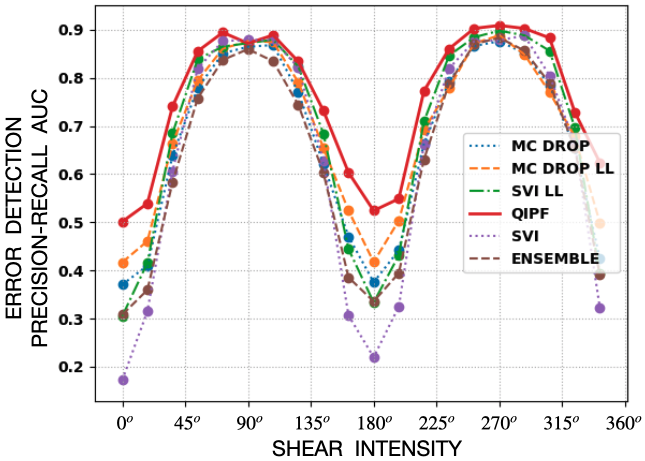}
    \caption{PR-AUC vs Shear Intensity}
  \end{subfigure}
  \begin{subfigure}{0.33\linewidth}
    \centering\includegraphics[scale = 0.2]{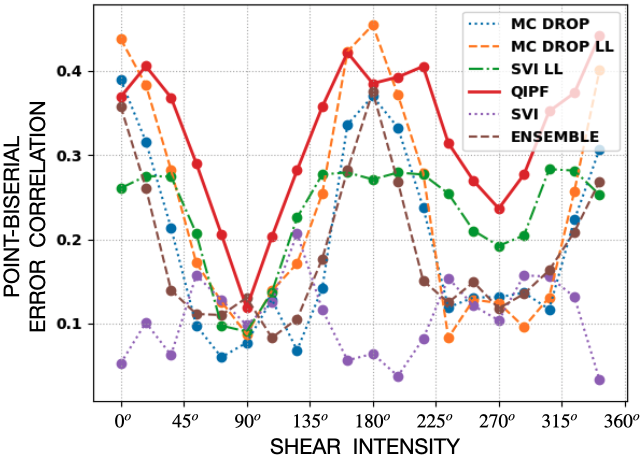}
    \caption{Error Corr. vs Shear Intensity}
  \end{subfigure}
  \caption{Results on CIFAR-10 VGG-3 model: (top row): Rotation Corruption, (bottom row): Shear Corruption, (left column): ROC-AUC vs noise intensity, (middle column): PR-AUC vs noise intensity, (right column): PT-biserial corr. vs noise intensity. QIPF can be seen to have significantly better performances than other methods.}
  \label{cif}
\end{figure*}

 \section{Experimental Results}
 We used python 3.7 along with tensorflow library to perform all simulations. We utilized MNIST, K-MNIST (Kuzushiji-MNIST, a more challenging alternative to MNIST) \citep{cla} and CIFAR-10 datasets for the comparative analysis of methods. For MNIST and K-MNIST, we used the regular 2-convolutional layer LeNet network architecture  with standard training procedure and dataset split. We used two different networks for modeling CIFAR-10, one of them being the same LeNet architecture and the other having a VGG-3 architecture consisting of 3 VGG convolution blocks (each consisting of two convolution layers followed by batch normalization, max pooling and dropout layers). The successive blocks consisted of 32, 64 and 128 filters respectively in each of their two layers. This was followed by two dense layers (with dropout in the middle) consisting of 128 and 10 nodes respectively. ReLu activation function was used here after each layer except the last, where Softmax was used. All networks were trained on uncorrupted datasets using Adam optimizer and categorical cross-entropy cost function. We achieved testing accuracies (on uncorrupted test-set) of 98.6\% for MNIST, 91.9\% for K-MNIST, 51.5\% for CIFAR-10 (using LeNet) and 88.5\% for CIFAR-10 (using VGG-3) thereby representing a diverse group of models on the basis of training quality, which is essential for evaluating the predictive UQ methods.\par
 
\begin{table*}[]
\centering
\resizebox{\textwidth}{!}{%
\begin{tabular}{|l|l|llllll|llllll|}
\hline
\multicolumn{1}{|c|}{\multirow{2}{*}{\begin{tabular}[c]{@{}c@{}}CORRUPTION\\ TYPE\end{tabular}}} & \multirow{2}{*}{DATASET}                                    & \multicolumn{6}{c|}{ROC-AUC}                                                                                                                                                                                                                         & \multicolumn{6}{c|}{PR-AUC}                                                                                                                                                                                                                          \\ \cline{3-14} 
\multicolumn{1}{|c|}{}                                                                           &                                                             & \multicolumn{1}{c}{MC-DROP}         & \multicolumn{1}{c}{MC-DROP-LL}      & \multicolumn{1}{c}{SVI}             & \multicolumn{1}{c}{SVI-LL}                   & \multicolumn{1}{c}{ENSEMBLE}        & \multicolumn{1}{c|}{QIPF}                     & \multicolumn{1}{c}{MC-DROP}         & \multicolumn{1}{c}{MC-DROP-LL}      & \multicolumn{1}{c}{SVI}             & \multicolumn{1}{c}{SVI-LL}                   & \multicolumn{1}{c}{ENSEMBLE}        & \multicolumn{1}{c|}{QIPF}                     \\ \hline
\multirow{4}{*}{ROTATION}                                                                        & MNIST                                                       & \multicolumn{1}{c}{0.63 $\pm$ 0.08} & \multicolumn{1}{c}{0.70 $\pm$ 0.09} & \multicolumn{1}{c}{0.55 $\pm$ 0.04} & \multicolumn{1}{c}{0.67 $\pm$ 0.06}          & \multicolumn{1}{c}{0.71 $\pm$ 0.09} & \multicolumn{1}{c|}{\textbf{0.74 $\pm$ 0.07}} & \multicolumn{1}{c}{0.65 $\pm$ 0.23} & \multicolumn{1}{c}{0.69 $\pm$ 0.20} & \multicolumn{1}{c}{0.60 $\pm$ 0.27} & \multicolumn{1}{c}{0.67 $\pm$ 0.24}          & \multicolumn{1}{c}{0.69 $\pm$ 0.21} & \multicolumn{1}{c|}{\textbf{0.73 $\pm$ 0.19}} \\
                                                                                                 & K-MNIST                                                     & 0.48 $\pm$ 0.05                     & 0.52 $\pm$ 0.09                     & 0.53 $\pm$ 0.03                     & 0.52 $\pm$ 0.08                              & 0.56 $\pm$ 0.11                     & \textbf{0.57 $\pm$ 0.12}                      & 0.77 $\pm$ 0.22                     & 0.79 $\pm$ 0.19                     & 0.79 $\pm$ 0.23                     & 0.79 $\pm$ 0.20                              & 0.81 $\pm$ 0.16                     & \textbf{0.84 $\pm$ 0.09}                      \\
                                                                                                 & \begin{tabular}[c]{@{}l@{}}CIFAR-10\\ (LENET)\end{tabular}  & \multicolumn{1}{c}{0.45 $\pm$ 0.02} & \multicolumn{1}{c}{0.45 $\pm$ 0.02} & \multicolumn{1}{c}{0.55 $\pm$ 0.01} & \multicolumn{1}{c}{\textbf{0.57 $\pm$ 0.03}} & \multicolumn{1}{c}{0.54 $\pm$ 0.03} & \multicolumn{1}{c|}{0.53 $\pm$ 0.02}          & \multicolumn{1}{c}{0.67 $\pm$ 0.11} & \multicolumn{1}{c}{0.68 $\pm$ 0.11} & \multicolumn{1}{c}{0.75 $\pm$ 0.09} & \multicolumn{1}{c}{\textbf{0.76 $\pm$ 0.08}} & \multicolumn{1}{c}{0.74 $\pm$ 0.08} & \multicolumn{1}{c|}{0.75 $\pm$ 0.11}          \\
                                                                                                 & \begin{tabular}[c]{@{}l@{}}CIFAR-10\\ (VGG-3)\end{tabular}  & \multicolumn{1}{c}{0.58 $\pm$ 0.07} & \multicolumn{1}{c}{0.61 $\pm$ 0.08} & \multicolumn{1}{c}{0.53 $\pm$ 0.02} & \multicolumn{1}{c}{0.63 $\pm$ 0.05}          & \multicolumn{1}{c}{0.57 $\pm$ 0.06} & \multicolumn{1}{c|}{\textbf{0.68 $\pm$ 0.07}} & \multicolumn{1}{c}{0.68 $\pm$ 0.12} & \multicolumn{1}{c}{0.69 $\pm$ 0.12} & \multicolumn{1}{c}{0.65 $\pm$ 0.18} & \multicolumn{1}{c}{0.72 $\pm$ 0.13}          & \multicolumn{1}{c}{0.66 $\pm$ 0.14} & \multicolumn{1}{c|}{\textbf{0.76 $\pm$ 0.09}} \\ \hline
\multirow{4}{*}{BRIGHTNESS}                                                                      & MNIST                                                       & 0.82 $\pm$ 0.11                     & 0.89 $\pm$ 0.08                     & 0.55 $\pm$ 0.05                     & 0.77 $\pm$ 0.02                              & \textbf{0.91 $\pm$ 0.06}            & 0.85 $\pm$ 0.04                               & 0.20 $\pm$ 0.12                     & 0.31 $\pm$ 0.10                     & 0.09 $\pm$ 0.13                     & 0.17 $\pm$ 0.19                              & \textbf{0.31 $\pm$ 0.15}            & 0.30 $\pm$ 0.14                               \\
                                                                                                 & K-MNIST                                                     & 0.57 $\pm$ 0.05                     & 0.75 $\pm$ 0.06                     & 0.53 $\pm$ 0.02                     & 0.73 $\pm$ 0.04                              & 0.86 $\pm$ 0.05                     & \textbf{0.90 $\pm$ 0.03}                      & 0.13 $\pm$ 0.05                     & 0.24 $\pm$ 0.05                     & 0.12 $\pm$ 0.07                     & 0.20 $\pm$ 0.06                              & 0.34 $\pm$ 0.04                     & \textbf{0.53 $\pm$ 0.05}                      \\
                                                                                                 & \begin{tabular}[c]{@{}l@{}}CIFAR-10\\ (LENET)\end{tabular}  & 0.43 $\pm$ 0.02                     & 0.46 $\pm$ 0.02                     & 0.53 $\pm$ 0.01                     & 0.58 $\pm$ 0.02                              & 0.56 $\pm$ 0.04                     & \textbf{0.64 $\pm$ 0.05}                      & 0.55 $\pm$ 0.15                     & 0.57 $\pm$ 0.13                     & 0.63 $\pm$ 0.13                     & 0.66 $\pm$ 0.11                              & 0.64 $\pm$ 0.11                     & \textbf{0.70 $\pm$ 0.08}                      \\
                                                                                                 & \begin{tabular}[c]{@{}l@{}}CIFAR-10\\ (VGG-3)\end{tabular}  & 0.69 $\pm$ 0.03                     & 0.75 $\pm$ 0.11                     & 0.52 $\pm$ 0.03                     & 0.69 $\pm$ 0.07                              & 0.69 $\pm$ 0.11                     & \textbf{0.80 $\pm$ 0.11}                      & 0.44 $\pm$ 0.16                     & 0.50 $\pm$ 0.16                     & 0.32 $\pm$ 0.24                     & 0.44 $\pm$ 0.19                              & 0.41 $\pm$ 0.19                     & \textbf{0.61 $\pm$ 0.12}                      \\ \hline
\multirow{4}{*}{SHEAR}                                                                           & MNIST                                                       & 0.62 $\pm$ 0.10                     & 0.68 $\pm$ 0.11                     & 0.56 $\pm$ 0.05                     & 0.62 $\pm$ 0.09                              & 0.69 $\pm$ 0.10                     & \textbf{0.70 $\pm$ 0.08}                      & 0.64 $\pm$ 0.22                     & 0.67 $\pm$ 0.22                     & 0.59 $\pm$ 0.26                     & 0.63 $\pm$ 0.25                              & 0.67 $\pm$ 0.22                     & \textbf{0.69 $\pm$ 0.21}                      \\
                                                                                                 & K-MNIST                                                     & 0.49 $\pm$ 0.03                     & 0.54 $\pm$ 0.08                     & 0.51 $\pm$ 0.03                     & 0.59 $\pm$ 0.06                              & 0.60 $\pm$ 0.12                     & \textbf{0.63 $\pm$ 0.13}                      & 0.65 $\pm$ 0.25                     & 0.68 $\pm$ 0.21                     & 0.67 $\pm$ 0.25                     & 0.71 $\pm$ 0.21                              & 0.72 $\pm$ 0.18                     & \textbf{0.77 $\pm$ 0.10}                      \\
                                                                                                 & \begin{tabular}[c]{@{}l@{}}CIFAR-10 \\ (LENET)\end{tabular} & 0.43 $\pm$ 0.01                     & 0.46 $\pm$ 0.01                     & 0.54 $\pm$ 0.01                     & 0.58 $\pm$ 0.01                              & 0.56 $\pm$ 0.02                     & \textbf{0.64 $\pm$ 0.02}                      & 0.54 $\pm$ 0.07                     & 0.57 $\pm$ 0.07                     & 0.63 $\pm$ 0.07                     & 0.66 $\pm$ 0.06                              & 0.63 $\pm$ 0.06                     & \textbf{0.69 $\pm$ 0.04}                      \\
                                                                                                 & \begin{tabular}[c]{@{}l@{}}CIFAR-10\\ (VGG-3)\end{tabular}  & 0.62 $\pm$ 0.07                     & 0.65 $\pm$ 0.08                     & 0.57 $\pm$ 0.03                     & 0.65 $\pm$ 0.04                              & 0.60 $\pm$ 0.17                     & \textbf{0.72 $\pm$ 0.06}                      & 0.66 $\pm$ 0.18                     & 0.69 $\pm$ 0.16                     & 0.63 $\pm$ 0.25                     & 0.68 $\pm$ 0.20                              & 0.64 $\pm$ 0.2                      & \textbf{0.75 $\pm$ 0.14}                      \\ \hline
\multirow{4}{*}{ZOOM}                                                                            & MNIST                                                       & 0.59 $\pm$ 0.15                     & 0.68 $\pm$ 0.16                     & 0.53 $\pm$ 0.05                     & 0.64 $\pm$ 0.09                              & 0.73 $\pm$ 0.14                     & \textbf{0.76 $\pm$ 0.13}                      & 0.36 $\pm$ 0.27                     & 0.43 $\pm$ 0.23                     & 0.34 $\pm$ 0.30                     & 0.40 $\pm$ 0.27                              & 0.45 $\pm$ 0.23                     & \textbf{0.55 $\pm$ 0.21}                      \\
                                                                                                 & K-MNIST                                                     & 0.50 $\pm$ 0.06                     & 0.59 $\pm$ 0.09                     & 0.52 $\pm$ 0.03                     & 0.63 $\pm$ 0.10                              & 0.70 $\pm$ 0.14                     & \textbf{0.72 $\pm$ 0.13}                      & 0.46 $\pm$ 0.28                     & 0.52 $\pm$ 0.24                     & 0.47 $\pm$ 0.29                     & 0.56 $\pm$ 0.23                              & 0.61 $\pm$ 0.19                     & \textbf{0.66 $\pm$ 0.16}                      \\
                                                                                                 & \begin{tabular}[c]{@{}l@{}}CIFAR-10\\ (LENET)\end{tabular}  & 0.43 $\pm$ 0.01                     & 0.46 $\pm$ 0.01                     & 0.54 $\pm$ 0.01                     & 0.57 $\pm$ 0.02                              & 0.56 $\pm$ 0.02                     & \textbf{0.64 $\pm$ 0.03}                      & 0.56 $\pm$ 0.11                     & 0.58 $\pm$ 0.10                     & 0.63 $\pm$ 0.10                     & 0.66 $\pm$ 0.08                              & 0.64 $\pm$ 0.09                     & \textbf{0.70 $\pm$ 0.06}                      \\
                                                                                                 & \begin{tabular}[c]{@{}l@{}}CIFAR-10\\ (VGG-3)\end{tabular}  & 0.60 $\pm$ 0.11                     & 0.63 $\pm$ 0.12                     & 0.52 $\pm$ 0.01                     & 0.59 $\pm$ 0.08                              & 0.61 $\pm$ 0.08                     & \textbf{0.67 $\pm$ 0.10}                      & 0.65 $\pm$ 0.19                     & 0.67 $\pm$ 0.17                     & 0.59 $\pm$ 0.26                     & 0.65 $\pm$ 0.21                              & 0.63 $\pm$ 0.22                     & \textbf{0.72 $\pm$ 0.15}                      \\ \hline
\end{tabular}%
}
\caption{Average ROC-AUC values (left) and average PR-AUC values (right) of different methods over all datasets and corruption types.}
\end{table*}

 To evaluate the UQ methods on the trained models, we corrupted the test-sets of the datasets with different intensities using benchmark corruption techniques \citep{hend}. For QIPF implementation, we first quantified the model PDF by evaluating the IPF (\ref{ipf}) of test-set raw predictions $\tilde{y}_k$ in the field formed by a downsampled set of training data raw predictions $\tilde{y}_i$ (where $i \in \{1...N\}$) so that $N = 6000$. The kernel width was set as the bandwidth determined by the Silverman's rule \citep{sil} multiplied by a factor that is determined through cross-validation over a part of the original/uncorrupted training dataset. This was followed by computation of the QIPF (\ref{sf}) and extraction of 4 QIPF uncertainty modes using the formulation in (\ref{vs}). We took the average value of the modes at any location to be the uncertainty score at that point. For other methods, we followed similar implementation strategy as \citet{laks2} and refer the reader to section A (appendix) for the details. We illustrate the QIPF implementation on MNIST in fig. 3. Fig 3a shows the IPF (estimated PDF) of the trained MNIST model, along with the locations of the test-set predictions on the IPF space, with the test-set being corrupted by 270 degrees of rotation. It can be observed here that in general, the corrupted test-set predictions are skewed towards the tail regions of the model PDF. The wrong predictions of the test-data (represented by red circles) can be seen to be more shifted towards the tail region of the model PDF than the correct predictions (green circles), thereby showing decent class-separation ability of the IPF between the wrong and correct prediction classes. The extracted QIPF uncertainty modes can be seen to markedly enhance the difference between the two classes with the first mode peaking almost exactly at the density peak of wrong predictions, while falling rapidly at the location of the correct predictions, that are more closer to model PDF mean. Fig 3b shows the histogram plots of uncertainty estimates of the different methods corresponding to the correct test-set predictions (blue) and the wrong predictions (red). The class-separation ability of the QIPF decomposition framework can be seen to be significantly better than the other methods with the most frequent uncertainty score values for the correct and wrong predictions being further apart from each other when compared to other methods. This is also evident from the ROC and precision-recall (PR) curves associated with prediction error detection of the different uncertainty quantification methods (i.e how accurately the uncertainty estimates detect model errors in real time) in fig. 3c and fig. 3d respectively where the QIPF performs much better than all of the other methods with the highest area under the curve (AUC) values. Fig. 3e and 3f show the ROC-AUC and PR-AUC associated with the UQ methods for different intensities/severity of rotation corruption. Also shown are the corresponding point-biserial correlation coefficients in fig. 3g between the test-set prediction errors and the uncertainty estimates at each of those intensities. One can observe here that the QIPF decomposition method dominates over all other methods in terms of error detection as well as error correlation. Fig. 4 shows the ROC-AUC vs corruption intensity (left column), PR-AUC vs corruption intensity (middle) and the point-biserial correlation vs corruption intensity (right) for the VGG-3 model trained on CIFAR-10 and with the test-set being corrupted by rotation corruption (top row) and shear corruption (bottom row). The QIPF decomposition framework can be seen here to perform significantly better than other methods. Table 1 summarizes the average values of ROC-AUC and PR-AUC for all models over the full range of corruption intensities for the different types of test-set corruption. The performance advantage of the QIPF decomposition framework can be seen to be generally consistent over all models and corruption types. We refer the reader to section B (appendix) for results on average point-biserial error correlation coefficients for all models over the full range of corruption intesities.
 
\section{Conclusion}
In this paper, we showed how a localized approach towards predictive uncertainty quantification yields much better results in terms of detecting predictive errors of a model under distorted test data. Specifically we showed how the Gaussian RKHS provides an efficient platform for local PDF estimation while also enabling the use of quantum physical operators to ultimately provide a fine-grained and high resolution description of uncertainty at each point in the model's prediction space. We plan to test the feasibility of this framework for other practical applications in deep learning and computer vision.





\bibliography{uai2021-template}

\appendix
\clearpage
\begin{table*}[!t]
\centering
\resizebox{\textwidth}{!}{%
\begin{tabular}{|l|l|llllll|}
\hline
\multicolumn{1}{|c|}{\multirow{2}{*}{\begin{tabular}[c]{@{}c@{}}CORRUPTION\\ TYPE\end{tabular}}} & \multirow{2}{*}{DATASET}                                   & \multicolumn{6}{c|}{POINT BISERIAL ERROR CORRELATION}                                                                                                                                                                                       \\ \cline{3-8} 
\multicolumn{1}{|c|}{}                                                                           &                                                            & MC-DROP                             & \multicolumn{1}{c}{MC-DROP-LL}      & \multicolumn{1}{c}{SVI}             & \multicolumn{1}{c}{SVI-LL}          & \multicolumn{1}{c}{ENSEMBLE}        & \multicolumn{1}{c|}{QIPF}                     \\ \hline
\multirow{3}{*}{ROTATION}                                                                        & MNIST                                                      & \multicolumn{1}{c}{0.16 $\pm$ 0.06} & \multicolumn{1}{c}{0.29 $\pm$ 0.06} & \multicolumn{1}{c}{0.07 $\pm$ 0.05} & \multicolumn{1}{c}{0.24 $\pm$ 0.04} & \multicolumn{1}{c}{0.31 $\pm$ 0.06} & \multicolumn{1}{c|}{\textbf{0.33 $\pm$ 0.09}} \\
                                                                                                 & K-MNIST                                                    & -0.01 $\pm$ 0.04                    & 0.02 $\pm$ 0.09                     & 0.03 $\pm$ 0.03                     & 0.03 $\pm$ 0.08                     & 0.08 $\pm$ 0.13                     & \textbf{0.09 $\pm$ 0.15}                      \\
                                                                                                 & \begin{tabular}[c]{@{}l@{}}CIFAR-10\\ (VGG-3)\end{tabular} & \multicolumn{1}{c}{0.14 $\pm$ 0.03} & \multicolumn{1}{c}{0.19 $\pm$ 0.08} & \multicolumn{1}{c}{0.05 $\pm$ 0.03} & \multicolumn{1}{c}{0.21 $\pm$ 0.06} & \multicolumn{1}{c}{0.13 $\pm$ 0.08} & \multicolumn{1}{c|}{\textbf{0.29 $\pm$ 0.08}} \\ \hline
\multirow{3}{*}{BRIGHTNESS}                                                                      & MNIST                                                      & 0.19 $\pm$ 0.02                     & 0.31 $\pm$ 0.02                     & 0.02 $\pm$ 0.03                     & 0.20 $\pm$ 0.10                     & \textbf{0.39 $\pm$ 0.06}            & 0.34 $\pm$ 0.05                               \\
                                                                                                 & K-MNIST                                                    & 0.06 $\pm$ 0.04                     & 0.25 $\pm$ 0.03                     & 0.04 $\pm$ 0.02                     & 0.23 $\pm$ 0.02                     & 0.42 $\pm$ 0.02                     & \textbf{0.49 $\pm$ 0.01}                      \\
                                                                                                 & \begin{tabular}[c]{@{}l@{}}CIFAR-10\\ (VGG-3)\end{tabular} & 0.26 $\pm$ 0.16                     & 0.35 $\pm$ 0.14                     & 0.03 $\pm$ 0.04                     & 0.23 $\pm$ 0.08                     & 0.26 $\pm$ 0.12                     & \textbf{0.34 $\pm$ 0.09}                      \\ \hline
\multirow{3}{*}{SHEAR}                                                                           & MNIST                                                      & 0.15 $\pm$ 0.12                     & 0.25 $\pm$ 0.11                     & 0.08 $\pm$ 0.04                     & 0.17 $\pm$ 0.11                     & \textbf{0.27 $\pm$ 0.10}            & 0.14 $\pm$ 0.07                               \\
                                                                                                 & K-MNIST                                                    & -0.01 $\pm$ 0.03                    & 0.06 $\pm$ 0.15                     & 0.02 $\pm$ 0.03                     & 0.13 $\pm$ 0.08                     & 0.17 $\pm$ 0.15                     & \textbf{0.19 $\pm$ 0.18}                      \\
                                                                                                 & \begin{tabular}[c]{@{}l@{}}CIFAR-10\\ (VGG-3)\end{tabular} & 0.19 $\pm$ 0.10                     & 0.24 $\pm$ 0.12                     & 0.10 $\pm$ 0.04                     & 0.23 $\pm$ 0.06                     & 0.18 $\pm$ 0.08                     & \textbf{0.32 $\pm$ 0.08}                      \\ \hline
\multirow{3}{*}{ZOOM}                                                                            & MNIST                                                      & 0.05 $\pm$ 0.10                     & 0.17 $\pm$ 0.12                     & 0.03 $\pm$ 0.03                     & 0.14 $\pm$ 0.08                     & 0.27 $\pm$ 0.13                     & \textbf{0.31 $\pm$ 0.13}                      \\
                                                                                                 & K-MNIST                                                    & 0.01 $\pm$ 0.03                     & 0.12 $\pm$ 0.19                     & 0.03 $\pm$ 0.03                     & 0.19 $\pm$ 0.13                     & 0.30 $\pm$ 0.19                     & \textbf{0.32 $\pm$ 0.18}                      \\
                                                                                                 & \begin{tabular}[c]{@{}l@{}}CIFAR-10\\ (VGG-3)\end{tabular} & 0.16 $\pm$ 0.16                     & 0.21 $\pm$ 0.17                     & 0.03 $\pm$ 0.03                     & 0.14 $\pm$ 0.10                     & 0.16 $\pm$ 0.10                     & \textbf{0.25 $\pm$ 0.14}                      \\ \hline
\end{tabular}%
}
\end{table*}

 \section{Implementation Strategy of Methods}
 Following are the implementation details of the methods we used for comparison:
 \subsection{Stochastic Variational Inference}
 For implementing the Stochastic Variational Inference algorithm \citep{blun}, we replaced each dense layer and the last convolutional layer in the network with Flipout layers. We used empirical Bayes for prior initialization of standard deviations and implemented negative log-likelihood Bayesian optimization to tune the initialization using 100 epochs.\par
 For implementing SVI-LL (last layer) algorithm, we replaced only the last dense layer with a Flipout layer. We implemented the same optimization procedure as before.
 
  \subsection{Monte-Carlo Dropout}
  For implementation of MC-Dropout, we introduced a dropout layer (during testing) after each convolutional block (before the next convolutional layer) and before each dense layer. We implemented 100 stochastic forward runs. The standard deviation of the results of all runs at each prediction was considered to be the uncertainty score. We varied the dropout rates between 0.1 and 0.4 and found that the best results were obtained by using a rate of 0.1 for LeNet models and 0.2 for VGG-3 model, which we used as benchmarks.\par
  
  The same implementation strategy was followed for MC-Dropout-LL (last layer), except that only a single dropout layer was introduced in the network before the last dense layer during testing. It was observed that the dropout rate of 0.2 corresponded to the best result for all models, which we used as benchmarks.
  
  \subsection{Ensemble}
  For implemeting the Ensemble methods, we trained 10 models with random initializations for each dataset and computed the standard deviation of results as the uncertainty score.
 
 \section{Error Correlation Results}
Average point-biserial error correlation coefficients between the model's test-set prediction errors and quantified uncertainties by the different techniques over all test-set corruption types and intensities are shown in table 2.


\end{document}